\documentclass[11pt]{article}
\usepackage{hyphenat}

\usepackage[]{acl}
\usepackage[T1]{fontenc}
\usepackage[utf8]{inputenc}
\usepackage{times}
\usepackage{latexsym}
\usepackage{microtype}
\usepackage{inconsolata}
\usepackage{url}
\usepackage{tikz}
\usetikzlibrary{positioning,arrows.meta}
\usepackage{comment}

\title{Looking under the Wrong Lamppost: On the Limitations of Automated Translation Quality Estimation
}

\author{
  Serge Gladkoff\\
  Logrus Global LLC
  \And
  Angelika Vaasa \\
  European Parliament, \\DG for Translation and Clear Language 
  \And
  Sue Ellen Wright \\
  Kent State University 
  \AND
  Ingemar Strandvik \\
  MQM Council
  \And
  Lifeng Han \\
  LUMC, Leiden \& Leiden University
}

\begin{document}
\maketitle

\begin{abstract}
Automation of Translation Quality Estimation (QE) has emerged as a widely discussed approach to managing translation quality at scale, and a growing number of tools and technologies have been released in pursuit of this goal. However, the proliferation of new QE systems has not always been accompanied by robust, transparent, and reproducible research and testing. This gap deserves critical scrutiny.
This paper examines some fundamental limitations of the QE technology from both theoretical and empirical perspectives, arguing that current QE systems are structurally ill-equipped to serve as reliable standalone tools in real-world translation workflows.
The reviewed evidence suggests that QE suffers from a range of interrelated and largely unresolved limitations. Most fundamentally, the evaluation of the quality of translation at the level of isolated segments is problematic because it tends to miss out on cohesion, coherence, and stylistic and rhetorical text features. In addition, empirical research documents several other limitations and flaws, including failure to generalize, systematic biases, overfitting and distribution collapse, performance gaps, error annotation challenges, and data scarcity. 
These are structural limitations arising from the complexity of human language and translation as a cognitive and communicative act - limitations that more data and better architectures have so far not overcome.

Consequently, segment-level QE scores should not be used as a standalone basis for routing, release, or review bypass in production; we argue future work should focus on automating human evaluation grounded in MQM.\footnote{To appear in the Proceedings of the 9th International Conference on Natural Language and Speech Processing (ICNLSP 2026), Trento, Italy, September 2026}

\end{abstract}

\section{Background}

The translation industry has long grappled with the misconception that translation is a task that anyone with basic language skills can perform. As a discipline within language science, translation studies have always faced the fact that language belongs to everyone because people use it daily for communication.

When a message is simple and clear, translation can indeed be a simple task. However, it can also require a high degree of sophistication. There is a long-standing body of work in translation studies that shows that certain linguistic and cultural phenomena are inherently resistant to direct translation \cite{catford1965linguistic,jakobson1959linguistic}.

Between these two extremes lies the full spectrum of content types and communicative modes that weave humanity into one interconnected world. Both realities coexist: some translation tasks are easy, while others are virtually unsolvable. This is precisely why relying solely on mechanical markers of quality, such as post-editing effort, fails to capture the full nuance of what translators do. As ~\newcite{yang2023rethinking} argue, word-level QE should be anchored in human semantic judgment rather than in the mere physical act of editing.

AI is helpful in many routine translation scenarios, even in crisis situations \cite{staiano-etal-2025-italert}, democratizing access to interlingual communication for a wide range of users. But AI also comes with a risk of capability misattribution: users perceive it as having capabilities that it does not possess. The shortcomings of machine translation (MT) and AI-assisted translation are not always self-evident. They are often detectable only by trained professionals who can identify and mitigate the significant and costly risks that arise from deploying unmanaged, unattended MT and AI in real-world settings.
As ~\newcite{mehandru2023physician} demonstrate in a medical context, even when aided by QE, non-professional users of AI, such as physicians, may over-rely on technology and miss critical errors that can lead to clinical harm.

At the same time, risk-averse real-world process management, as mandated by ISO 9000, presumes that production workflows have been tuned to avoid even relatively rare risk events.\footnote{\url{https://www.iso.org/standards/popular/iso-9000-family} specifically ISO 9001:2015} Applying the same logic to translation production workflows means that MT and AI must be incorporated sensibly to achieve manufacturing-grade reliability, risk guarantees, and robustness. A QE approach that focuses exclusively on pre-production possibilities rather than post-production realities is difficult to reconcile with this requirement, as reflected by the findings of \newcite{scarton-etal-2019-estimating}, and \newcite{liu-etal-2025-introducing}.

With the advent of MT and AI translation, the industry has gained powerful new tools, but these tools are far from perfect, even for language pairs where English is one of the languages, let alone for other language pairs \cite{han2024neural}. 
Researchers and practitioners increasingly recognize that the language problem is far from being solved \cite{liangmetahope}. Therefore, the industry needs to adopt a more rigorous and critical approach, one that prioritizes factual, reliable, and reproducible research.

\section{Fundamental Flaw of the Segment-Based Approach}

As research approaches the final, and arguably most difficult, frontier of language automation, it is important not to be drawn in by its promise alone, but to focus on where genuine solutions are most likely to emerge. Current research efforts may be disproportionately concentrated on areas that are computationally convenient rather than theoretically appropriate, or even pragmatically accurate.

Broad claims are made about the promise of QE systems. However, in professional, operational environments,
the path from theoretical promise to real-world deployment requires rigorous, transparent, and reproducible experimental evidence to establish both safety and efficacy. In the absence of such evidence, or where evidence points to the contrary, these claims should be treated with skepticism.

This situation evokes the classic streetlight effect—an observational fallacy where a person searches for lost keys under a lamppost simply because the light is better there, rather than in the dark alley where the keys were actually dropped. In automated evaluation, focusing exclusively on isolated, segment-level metrics represents searching where computation is easy and convenient, rather than venturing into the darker, more complex domain of extra-segmental context, linguistic multidimensionality, discourse structure, human perception, and communicative intent where true translation quality actually resides.

The notion of fully automated QE appears to conflict with fundamental properties of language, a fact that may be intuitive to professional translators or scholars of translation studies, but one that can also be demonstrated through simple, measurable analysis. 
Statistical and mathematical considerations suggest that assessing the translation quality of a single isolated sentence, without its contextual environment, is problematic \cite{gladkoff2022uncertainty}. This principle is also reflected in the Multi-Range Theory of Translation Quality Measurement advanced by the MQM Council \cite{lommel2024multirange}.

Constructing metrics to reliably represent translation quality at the level of isolated sentences is problematic because translation correctness depends strongly on extra-segmental context. A translation of a sentence can be simultaneously correct and incorrect depending on context, terminology settings, quality requirements, target audience, author intent, or source meaning that cannot be recovered from the sentence alone. In other words, extra-segmental factors are determinative. Even spelling and grammatical errors cannot always be reliably identified from a single segment in isolation.

Serious translation errors are distributed across the full range of segment-level QE scores \cite{gladkoff2025hardfact}, as shown on Figure 1.
Consequently, those scores cannot serve in the production environment as a reliable basis for routing, auto-release, review bypass or prioritization of human review effort, because they do not provide a stable, monotonic ordering of risk. At best, such scores may be retained for retrospective research or descriptive analytics, but should not serve as decision variables in production.

\begin{figure*}[t]
\centering
\includegraphics[width=\textwidth]{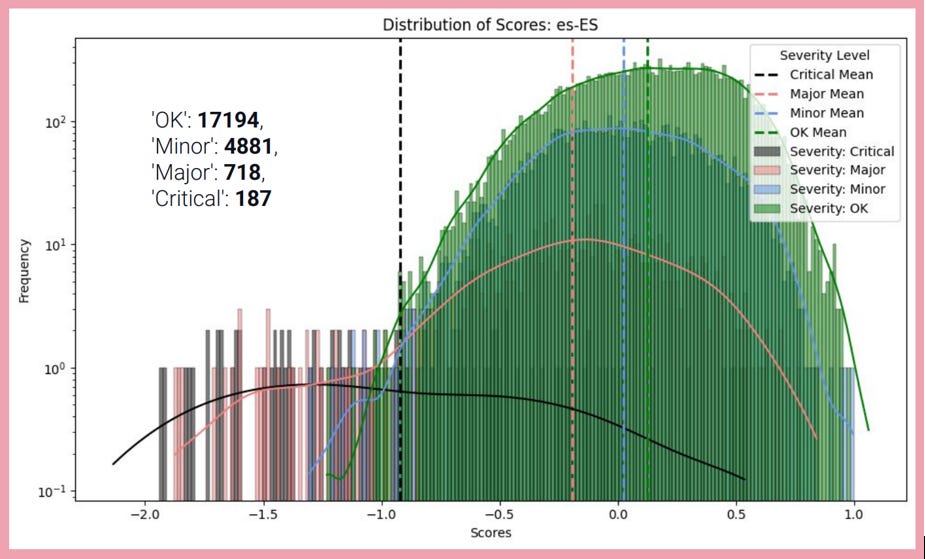}
\caption{Automatic translation quality prediction scores vs. actual error distributions. (COMET model wmt20-comet-qe-da). Note that Y-axis is in log scale.}
\label{fig:revix-aggregate}
\end{figure*}

Proponents of segment-based evaluation sometimes argue that many contextual aspects can be derived from a training corpus. We argue that this is a misconception and one with a longer history in translation technology. Before the advent of CAT tools, translators translated thoughts, not sentences. When translation memories arrived, many translators objected to the idea of dissecting a text into constituent parts and then claiming a productivity gain by reassembling new content from old fragments. Their concern was that context was being lost and that even 100\% translation-memory matches might require different translations in a new context \cite{dragsted2004segmentation,ranasinghe2020intelligent}. 
If we consider what translators actually do when working in CAT tools, we see that they are not merely verifying translation-memory matches. They are reconstructing the meaning of source content using fragments of previous translations stored segment by segment and stripped of their original contextual reference.

From a formal perspective, translation quality can be viewed as a function not only of the source and target segments, but also of contextual variables, such as discourse structure, terminology constraints, and communicative intent. Because these variables are not recoverable from isolated segments, any segment-level QE necessarily operates under incomplete information, which limits its validity as a standalone measure.

Translation quality could be conceptualized as a function $Q(s,t,c)$, where $s$ denotes the source segment, $t$ the target translation, and $c$ the set of contextual parameters relevant to specifications-based quality assessment.

\paragraph{Linguistic Example:}
Consider the Spanish source segment "Tiene razón" and its translation into target languages that require grammatical gender marking for pronouns used to resolve the autonomous Spanish verb forms. Whether a particular translation is correct depends on discourse context establishing the referent's gender ("he/she is right"), communicative register (if the register is formal, this could mean "you are right"), or whether it refers to an entity (e.g. the Government, "it is right"). Without access to sufficient contextual information, multiple distinct translations may each be plausible, and no sentence-level QE model can determine which variant is appropriate based on the segment alone.
More generally, identical source–translation segment pairs may receive different quality judgments depending on contextual variables external to the segment itself, illustrating that sentence-level translation quality is not uniquely determined by isolated segment pairs.



\paragraph{Terminology/Domain Example:}
Consider the source segment “Apply the patch.” In a software documentation context, patch may refer to a software update, whereas in a medical context, it may refer to a transdermal medication patch. A translation that is correct in one domain may be incorrect in another, even if the isolated segment pair remains identical. Without domain context, sentence-level QE cannot reliably determine translation correctness.

Therefore, without domain context, segment-level QE is forced to evaluate translations against moving targets.

\section{
Limitations of Current QE Systems}
\label{new-sec-merging-limitations}

\subsection{Generalization and Distribution Shift}

{\textit{Overfitting and Distribution Collapse}.}
When a QE model is trained on a corpus of past translations, it does not learn how to translate from this corpus. Rather, it learns the patterns of translations deemed correct within the available data, inevitably diluted by translation errors stored in the corpus itself.

Even when a QE model is trained on an exceptionally clean translation corpus, and even when training yields apparently strong results on in-domain test data, applying the model outside the distribution of its training corpus can lead to substantial degradation in predictive reliability. A recent benchmark for English--Hebrew QE directly confirms the instability of QE models when applied outside their training distributions. ~\newcite{rosenbaum2026mtqeenhe} introduced an MTQE.en-he benchmark and documented fine-tuning experiments using widely adopted models such as TransQuest and COMETKiwi. Their results exhibited sensitivity to overfitting and distribution collapse. This finding is particularly significant because it arises even with clean annotated data and pre-trained models backed by substantial computational resources.

This evidence suggests that even state-of-the-art QE systems cannot reliably generalize beyond their specific training contexts, validating long-standing concerns about QE instability in real-world scenarios characterized by linguistic and use-case diversity. 
Morphological complexity, a hallmark of many languages, appears to be a particularly challenging dimension for current QE models to navigate, suggesting that existing training procedures have limited ability to learn language-agnostic quality indicators \cite{chahuneau2013translating,siani2026semisyntheticparalleldatatranslation}.

The limitation appears to hold even when the training corpus is very large. 
\newcite{gladkoff2025hardfact} reports on a large-scale, real-world case study by Bane covering technical content in 9 languages, using an industry standard pre-trained QE model. The study found that segments containing major and critical errors received scores spanning the entire range of the model's output—including the highest scores. In more than one language, critical errors appeared even among the top-scoring 5 sentence pairs out of a set of 2,000. 
When serious errors are distributed across the full score range in this way, the QE score becomes difficult to use in practice, as no threshold can reliably separate translations that are fit for use as-is from those that require human review (see Figure 1).

These limitations help explain why the field of translation QE continues to encounter significant challenges with respect to the robustness and stability of automatic evaluation systems, particularly at the segment and span levels. The findings of the WMT25 Shared Task also highlighted persistent performance gaps consistent with these larger concerns \cite{lavie2025wmt25}.
The same kind of limitations at the model level are reflected in the performance of downstream evaluation, as discussed below.

\noindent {\textit{Domain Variation and Generalization Failures}.}
One of the most critical vulnerabilities of QE systems is their inability to generalize across domains and data sources. Large cross-domain variations in auto-rater behavior are supported by multiple studies that document systematic domain generalization failures.
%
~\newcite{kocyigit2022better} demonstrate this failure in parallel corpus mining settings, where models trained on specific datasets could not reliably assess quality in broader real-world contexts.
~\newcite{sharami2023tailoring} show that even domain-tailored models struggle to maintain robustness when confronted with text characteristics or language pairs outside their adaptation scope.

\noindent{\textit{Synthetic Data Generation Issues}.}
The reliance on synthetic data -- machine-generated translations with pseudo-labels---for training QE systems introduces additional stability concerns. ~\newcite{geng2025alleviating} show that pseudo-labels frequently fail to align with human preferences. The distribution shift between synthetic pseudo-translations and real-world translations creates systematic mismatches that undermine model generalization.

The reliance on massive parallel corpora also creates what ~\newcite{kocyigit2022better} describe as a structural feedback loop in which only organizations with large existing data resources can build the tools necessary to find more data. Current QE models therefore risk becoming over-optimized for the narrow distributions represented in their training sets.

Because high-quality 
human-annotated translation quality data
are scarce and expensive, synthetic data is difficult to avoid in practice. However, this reliance introduces exactly the type of distribution shift that makes robust deployment difficult.

\subsection{Annotation and Human–Machine Misalignment}
\label{new-subsec_human-machine-misalign}
\noindent {\textit{Human--Machine Misalignment}.}
A fundamental issue undermining QE stability is the misalignment between what human annotators label as errors and what QE systems learn to predict. ~\newcite{yang2023rethinking} find that traditional post-editing effort labels often conflict with human judgment about whether a word is actually well or poorly translated. This suggests that QE systems trained on conventional datasets may be optimizing for proxy metrics that do not closely correspond to true translation quality.

The problem is compounded by a misalignment between model targets and human workflow needs. As \newcite{sarti2025qe4pe} demonstrate, traditional QE labels often conflict with post-editors' actual preferences. Thus, even when QE models achieve high correlation scores against standard benchmarks, their predictions offer limited practical utility because they fail to capture what human editors genuinely need to correct.

This gap between human judgment and automatic systems is particularly pronounced at the word and span levels. \newcite{sarti2025annotators} show that relying on single-annotator evaluations masks underlying instability; when multiple sets of human labels are introduced, supervised methods struggle with severe label uncertainty. In classical terms, this highlights a fundamental issue of inter-rater reliability in fine-grained QE annotation \cite{scarton-etal-2019-estimating}.
\newcite{mehandru2023physician} also show in a high-stakes medical setting that QE can help identify some translation issues, but it still fails to reliably detect critical errors compared to more manual methods, such as back-translation.

Research on fine-grained error prediction confirms that identifying exact error boundaries and assessing severity remains highly challenging.

Taken together, these findings suggest that the usability of QE in professional post-editing workflows remains substantially limited. Domain, language pair, and editor expertise all emerge as critical factors in determining QE effectiveness, indicating that current QE systems are highly context-dependent, brittle, and unreliable.

\subsection{Biases and Error Detection Limitations}

\noindent {\textit{Performance Gaps and LLM-based Auto-Rater Limitations}}. Recent work has shown that LLM-based auto-raters (sometimes called “LLM-as-a-judge”, although neither term defines the exact method precisely) typically underperform on fine-grained quality assessment tasks. The difficulty of identifying fine-grained error spans and accurately classifying their severity represents a fundamental limitation of current QE approaches. ~\newcite{jung2023enhancing} show that even when the task is decomposed into sequential stages --sentence-level detection followed by word-level span extraction---the individual components continue to struggle with accuracy. Studies on Chinese homophone translation also indicate that while larger LLMs exhibit greater stability and robustness to perturbations, they still show substantial vulnerabilities when evaluating nuanced semantic content \newcite{qian2025homophone}.

A major obstacle to the adoption of QE in production is its opaque nature. As ~\newcite{dinh2023perturbation} argue, most current QE systems fail to provide explainable feedback, making it difficult for human editors to trust or verify their outputs. Their perturbation-based QE framework points toward a possible shift toward unsupervised, explainable, and transparent methods that do not rely as heavily on biased training corpora, although it also illustrates how attempts to solve one set of QE problems can introduce additional complexity.

\noindent {\textit{Error Detection Limitations}.}
Auto-raters continue to struggle with precise error detection, span annotation, and severity classification, falling short of the precision required by professional MQM standards \cite{jung2023enhancing}.

Extensive analysis of recent WMT shared tasks indicates that, while sentence-level QE has improved considerably, finer-grained tasks such as word-level QE and error span detection remain a challenge. ~\newcite{qian2025homophone} show that larger LLMs are somewhat more stable, but still vulnerable when evaluating translations involving \textit{nuanced semantic content}. Likewise, although xCOMET represents a substantial step toward transparency, practical implementation of such granular schemes still falls short of the MQM gold standard for reliably distinguishing error types and severities \cite{guerreiro2023xcomet}.

\noindent{\textit{Systematic Biases}.}
Recent research has exposed systematic biases embedded in QE metrics themselves. ~\newcite{zhang2025penalizing} show that QE systems consistently assign lower quality scores to translations as they become longer, even when those translations are entirely accurate. This length bias arises because training data is skewed: models learn to associate length with error probability and therefore penalize paraphrastic translations, regardless of whether they are perfectly fit for their purpose.

Another documented bias concerns gender. ~\newcite{zaranis2025watching} show that, when gender is ambiguous in the source text, current QE models systematically favor masculine inflections with higher quality scores, regardless of actual translation accuracy. These findings suggest that QE metrics may conflate translation quality with superficial patterns in skewed training data.

\noindent{\textit{Cross-System Heterogeneity and Inconsistency}.}
Applying QE metrics to outputs from heterogeneous MT systems reveals further inconsistencies. The same QE model can produce markedly different results depending on the MT system being evaluated, suggesting that QE models may learn system-specific regularities rather than generalizable quality indicators.

\newcite{dinh2023perturbation} partly addresses this issue through an explainable perturbation-based approach designed for black-box MT. Their work is valuable, but it also illustrates a broader challenge: QE often introduces new layers of complexity in the course of trying to solve older ones.

\section{Limited Assistance, No Routing Validation for Segment-Level QE}
\label{sec-relocate-Limited_assis_No_Routing}
The literature on sentence-level QE can be broadly divided into two distinct lines of inquiry: (i) whether QE information can assist human translators in constrained MT post-editing scenarios, and (ii) whether sentence-level QE scores can serve as reliable standalone signals for production routing decisions, such as determining which segments may bypass human review. These two questions are frequently conflated, although existing evidence does not support treating them as equivalent.

On the first question, the evidence is limited and mixed. Some studies report partial or conditional benefits when QE information is shown to human post-editors. ~\newcite{huang2014adaptive} report productivity gains in a document-specific post-editing setting. ~\newcite{turchi2015does} find only slight gains overall, with statistical significance limited to restricted conditions of segment length and MT quality. ~\newcite{bechara2021role} report lower average post-editing time and effort when MT QE information was provided, but their study is relatively small. More recently, ~\newcite{liu-etal-2025-introducing} concluded that empirical evidence for the usefulness of sentence-level QE remains limited and inconsistent. Importantly, these studies are about human-in-the-loop assistance, not autonomous acceptance, rejection, or routing of segments.
Integrated CAT-tool studies such as IntelliCAT \cite{lee2021intellicat} report productivity improvements, but because sentence-level QE is bundled with other assistive features, the contribution of QE alone cannot be isolated.

To our knowledge, the main published attempt to validate sentence-level QE as an explicit routing mechanism is the APE-QUEST case study 
~\newcite{alva2021validating}. This study reports that, in a narrowly defined public-administration workflow for English--Dutch and English--French, thresholded QE scores could reduce time and cost while preserving roughly similar end-user acceptability. However, these results should not be interpreted as a general validation of sentence-level routing. It is a single case study restricted to two language pairs, one domain, and two specific use cases; it depends on threshold tuning for each scenario; and the authors themselves show that gains are larger with oracle scores than with predicted ones.

By contrast, the counter-evidence is more directly relevant to the production-routing claim. ~\newcite{zaretskaya2020estimation} find that an industry-level QE system was not reliable enough for the selection of MT engines when candidate systems had similar quality. ~\newcite{glushkova2021uncertainty} argue that segment-level point estimates provide limited information and are often unreliable because they are trained on noisy, biased, and scarce human judgments. Most recently, the WMT25 Shared Task confirmed that segment level prediction remains difficult, reference-based baselines still outperform LLMs at the segment level, accurate error detection remains challenging, and even quality-informed minimal correction is difficult across diverse languages \cite{lavie2025wmt25}.

The strongest production-oriented negative evidence is the large-scale industry case study reported by \newcite{gladkoff2025hardfact}. In that work, severe and critical errors overlap broadly across the sentence-level QE score range, so that no reliable threshold can separate safe-to-pass segments from those requiring review. If serious errors are distributed across the full score continuum, then the score fails not only as a cut-off variable, but also as a ranking variable to prioritize the review effort (see~Figure1).
Consequently, the current literature does not provide robust and replicated support for the use of raw segment-level QE scores as standalone production routing criteria. At most, it provides limited and mixed evidence that QE information may sometimes assist human post-editors in narrowly defined workflows. That is a materially weaker claim, and it does not support broader claims in the literature and practitioner discourse that segment-level QE is ready for generalized no-touch routing in production.

\section{Empirical Validation with a Large Dataset}
\label{sec:empirical-validation}

The empirical test from large industrial data reported in this section was designed to address the operational question that follows from the preceding discussion. The question is not simply whether a segment-level QE system can produce a score that is statistically different from random. The operational question is stricter: 
can that score be used as a reliable proxy for translation quality for routing, bypassing review, or classification decisions. In particular, we tested whether a single commercial segment-level QE score could be treated as a calibrated quality score, as a portable threshold, or at least as a reliable ranking signal for the prioritization of segments for review.

\paragraph{Data and the gold standard.}
The experiment used three MQM-annotated data sources representing different quality regimes. One data set was from the German Research Center for Artificial Intelligence (DFKI). The DFKI material consisted of legacy rule-based machine translation output with errors in nearly every sentence. The second dataset consisted of MT-generated, unedited translations from the European Parliament (EP). The third consisted of final human-edited translations from the European Commission (EC), representing a low-error production-quality regime. These three sources are important because they span the practical range in which QE is normally claimed to be useful: very poor MT output, an unedited AI/MT output from an engine considered good, and high-quality human-edited final translations.

Across the three primary corpora, the analysis covered \textbf{104,762 scored segments}: 2,543 DFKI segments, 56,715 EP segments, and 45,504 EC segments. All segments had undergone human MQM annotation, which was used as the gold standard. The main binary gold label was whether a segment contained any MQM error. A secondary major-plus AUC was also recorded where available, but the thresholded confusion matrices were generated for the any-error condition. This distinction matters because major-error counts were often sparse in the EC slices, making major-plus-threshold conclusions unstable.

\paragraph{QE model and analysis instrument.}

The segments were scored using the TAUS EPIC QE model. The resulting scores and MQM annotations were analyzed using Revix, a statistical analysis platform designed for the evaluation of translation-quality workflows. Revix stores annotated translation datasets and provides statistical analyses of the relationship between QE outputs and human quality annotations, including ranking performance, calibration, threshold behavior, and workflow-level utility metrics. In this study, Revix was used solely as an analysis environment; the object of evaluation was the TAUS EPIC QE score rather than the Revix platform itself.

Since TAUS EPIC QE reports a quality-oriented score, we treated lower TAUS scores as higher predicted error risk and evaluated the review order from the lowest TAUS score upward. 

We then evaluated the score as a decision instrument, not merely as a correlation number. 
For each corpus/language-pair slice, we computed ROC AUC, AUPRC and AUPRC lift over random selection, precision, recall, specificity and $F$-scores at candidate thresholds, confusion matrices, triage-gain curves, calibration curves, Brier score relative to a no-skill base-rate predictor, bootstrap confidence intervals, and permutation p-values.

The Revix export contained 186 snapshots. Each snapshot corresponds to a corpus/language-pair analysis under a specific threshold strategy and contains the resulting ranking, calibration, threshold, and workflow-utility statistics. 
Because several snapshots represented the same corpus/language-pair slice under different threshold strategies, the corpus-level analysis duplicated them into 72 unique dataset/language-pair slices. 
The threshold-specific analysis retained the full set of 186 snapshots, because the threshold choice is itself part of the operational decision. The saved threshold scenarios included maximum $F_1$, maximum $F_2$, maximum $F_{0.5}$, recall-at-least-90\%, maximum Youden's $J$, maximum triage gain, and manual or unspecified thresholds.

\begin{table*}[t]
\centering
\small
\begin{tabular}{lrrrrrrr}
\hline
Corpus & Slices & N & Base & AUC & Prec.@$F_1$ & Rec.@$F_1$ & Brier/no-skill \\
\hline
DFKI & 2 & 2,543 & 93.3\% & 0.679 & 93.5\% & 100.0\% & 11.139 \\
EC & 20 & 45,504 & 6.5\% & 0.591 & 9.4\% & 85.8\% & 1.756 \\
EP & 23 & 56,715 & 32.5\% & 0.631 & 40.9\% & 95.5\% & 1.404 \\
\hline
\end{tabular}
\caption{Aggregate Revix results by primary corpus. Base rate is the share of segments with at least one MQM error. Precision and recall are reported at the threshold that maximizes $F_1$ in each slice. A Brier/no-skill ratio above 1.0 means that the QE score is worse calibrated than a constant predictor that always returns the corpus base rate.}
\label{tab:revix-aggregate}
\end{table*}

\paragraph{Aggregate finding.}
Table~\ref{tab:revix-aggregate} shows that TAUS EPIC QE contains some statistical signal. The median AUC is above random for EC and EP, and many p-values are significant because the number of segments is large. However, the effect size is modest. The median EC AUC is only 0.591 and the median EP AUC is only 0.631. These values are compatible with weak-to-moderate ranking, not with reliable production classification.

The calibration results were even poorer.
In every calibrated unique slice in the export, the Brier score was worse than the no-skill base-rate baseline; the Brier/no-skill ratio was above 1.0 for all 70 slices where calibration statistics were reported. This means that the score should not be interpreted as a probability-like estimate of translation quality or error risk. Even when the score is somewhat correlated with the presence of errors, it is not calibrated enough to support an absolute quality interpretation. This directly undermines the idea that the output is a general \textit{translation quality score}.

\paragraph{Corpus-level behavior.}
The DFKI corpus illustrates why a statistically positive signal is not necessarily useful. Its error base rate was approximately 90--96\%. In such a corpus, binary error/no-error triage is almost meaningless because nearly every segment is erroneous. The strongest AUC in the export occurred in this high-error regime, but that does not imply useful pass/fail triage. It means only that the score can sometimes rank very bad material ahead of less bad material in a corpus where almost everything already requires attention.

The EC corpus represents the opposite case. Its median base rate for any error was only 6.5\%. In such a low-error regime, even a non-random ranker struggles to produce useful precision. For example, among the better EC slices, English--Greek reached AUC 0.692 but only 9.5\% precision at peak $F_1$; English--Spanish reached AUC 0.662 but only 11.1\% precision; English--Italian reached AUC 0.626 but only 7.3\% precision. These results do not yield operationally persuasive targeted review lists. They can enrich an audit sample relative to random selection, but most flagged segments are still clean. For EC-like production-quality files, the evidence supports at most audit enrichment after local validation, not autonomous triage.

The EP corpus is the one where TAUS EPIC QE looks most plausible, because the base error rate is higher. The median base rate for any-error was 32.5\%. The selected language pairs had moderate rank-order behavior: English--Czech reached AUC 0.695 with 48.2\% precision at peak $F_1$; English--Italian reached AUC 0.673 with 35.4\% precision; English--Polish reached AUC 0.670 with 42.0\% precision; English--Dutch reached AUC 0.665 with 54.3\% precision; English--Spanish reached AUC 0.664 with 44.5\% precision. These figures show that the model can sometimes place more erroneous segments earlier for review. However, the weaker EP slices remained close to random for practical purposes: Irish, Latvian, Swedish, Estonian, Slovenian, and Hungarian all had AUCs below 0.60. Even in the best EP cases, the evidence can support the prioritization of segments for review, but not a safe threshold for bypassing the review.

\paragraph{Threshold behavior and triage.}
A central empirical result is that the operational verdict strongly depends on the threshold policy. Table~\ref{tab:revix-thresholds} summarizes two important threshold strategies. The maximum-$F_2$ strategy emphasizes recall, while the maximum yield strategy emphasizes savings in the review effort.

\begin{table*}[t]
\centering
\small
\begin{tabular}{llrrrrrr}
\hline
Corpus & Threshold strategy & Snapshots & Median $t$ & Flagged share & Precision & Recall & RCR-MEC \\
\hline
EC & Max $F_2$ & 20 & 0.908 & 70.6\% & 9.1\% & 93.7\% & 6.9\% \\
EC & Max triage gain & 18 & 0.878 & 42.7\% & 11.3\% & 60.5\% & 44.4\% \\
EP & Max $F_2$ & 19 & 0.951 & 80.8\% & 39.8\% & 99.0\% & $-4.6$\% \\
EP & Max triage gain & 23 & 0.891 & 42.0\% & 45.4\% & 60.4\% & 39.3\% \\
\hline
\end{tabular}
\caption{Threshold behavior for EC and EP. RCR-MEC stands for "Modeled review-cost reduction, excluding missed-error cost": the value estimates the reduction in QE-plus-human-review expenditure relative to full human review. It excludes the operational, corrective, and risk costs of errors remaining in unreviewed segments. Under the maximum-triage-gain condition, approximately 40\% of error-containing segments were not captured.}

\label{tab:revix-thresholds}
\end{table*}

The trade-off is clear. High recall usually requires flagging most of the segments in a file. In EP data, the maximum-$F_2$ threshold catches almost all errors, but it flags a median of 80.8\% of segments and has a negative median saving proxy (RCR-MEC, Modeled review-cost reduction, excluding missed-error cost). In EC data, the same strategy indicates a median of 70.6\% of segments while producing only 9.1\% precision. In contrast, the maximum-triage-gain strategy saves review effort, but it captures only about 60\% of errors in both EC and EP datasets. Thus, the apparent gain is achieved by leaving a substantial share of errors in the unreviewed tail.

We also applied a stricter operational classification screen: a threshold would have to review at most 50\% of the file, catch at least 80\% of actual errors, and make the flagged set at least $1.5\times$ richer in errors than random selection. No saved threshold scenario in the export satisfied all three conditions. This screen provides the clearest empirical result for production use. 

The key operational finding is: If triage is defined as materially reducing review effort while catching the vast majority of errors, the export data does not support TAUS EPIC QE as a safe triage mechanism.

\paragraph{Ranking is not triage.}
The results show the importance of distinguishing ranking from triage. A ranker answers the question: which segments should a reviewer inspect first if review capacity is limited? A triage gate answers a stronger question: which segments can safely bypass review? TAUS EPIC QE sometimes provides a weak answer to the first question. It does not provide a reliable answer to the second one. A score that enriches the first part of a review list is not automatically a score that can justify leaving the rest unchecked.

\begin{figure}[t]
\centering

\begin{tikzpicture}[
    node distance=.3cm and .3cm,
    box/.style={
        draw,
        rounded corners,
        align=center,
        minimum width=3.8cm,
        minimum height=0.9cm
    },
    >={Latex[length=3mm]}
]

\node[box] (l1) {Same QE scores};

\node[box, below=of l1] (l2)
{Threshold A\\(Ranking objective)};

\node[box, below=of l2] (l3)
{Useful ranker};

\draw[->] (l1) -- (l2);
\draw[->] (l2) -- (l3);

\node[box, right=of l1] (r1)
{Same QE scores};

\node[box, below=of r1] (r2)
{Threshold B\\(Workflow objective)};

\node[box, below=of r2] (r3)
{Unsafe routing signal};

\draw[->] (r1) -- (r2);
\draw[->] (r2) -- (r3);
\end{tikzpicture}
\caption{
Identical QE scores can lead to different conclusions depending on the evaluation objective. A threshold selected to maximize ranking performance may yield a useful ranker, whereas a threshold selected to maximize workflow utility may reveal that the same scores are unsuitable as a routing signal. This illustrates that ranking quality and routing utility are not equivalent.
}
\label{fig:ranking-vs-routing}
\end{figure}
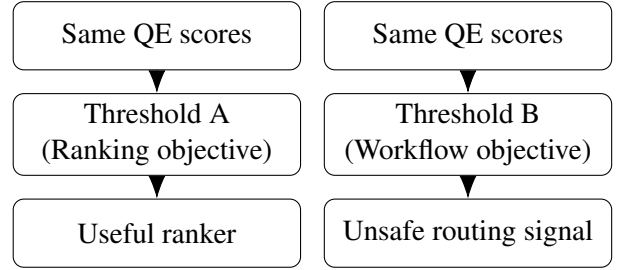

The distinction between ranking performance and workflow utility is critical. A QE model may produce a statistically useful ranking while still failing to generate safe operational decisions when deployed under a specific threshold strategy. The same underlying score distribution can therefore lead to substantially different conclusions depending on the evaluation objective, as illustrated in Figure \ref{fig:ranking-vs-routing}. 







This distinction explains why the same corpus/language-pair slice could receive a ``useful ranker'' verdict under one threshold policy and a negative workflow verdict under another.
Among the 72 unique slices, 50 had more than one saved verdict in the threshold scenarios. This is not just noise in the reports. It shows that usability is not an intrinsic property of the QE model alone. It is a property of the model, the corpus, the language pair, the error base rate, the threshold, and the relative cost of making different translation mistakes. Any claim that a QE model is ``usable'' is therefore insufficient unless it specifies the operating condition under which it is usable.

\paragraph{Non-portability across corpus conditions.}
The same target language did not behave consistently across EC and EP datasets. Across the 20 common target languages, the correlation between EC AUC and EP AUC was only 0.379. For English--Czech, AUC was 0.583 in EC but 0.695 in EP. For English--Swedish, AUC was 0.488 in EC and 0.586 in EP. In contrast, English--Portuguese and English--Greek were stronger in EC than in EP. The practical precision changed even more sharply because the base rates differed: a modest AUC in EC often produced a very low absolute error rate in the flagged set, while a similar AUC in EP produced a more persuasive targeted review list because the underlying error rate was much higher.

This finding is important for deployment. A threshold or language-pair conclusion reached on one dataset cannot be assumed to be transferred to another file. The language pair alone is not enough. Domain, data source, MT system, editorial status, and expected error base rate all affect whether the QE score has any operational value.

\paragraph{Interim conclusions.}
The empirical results support four interim conclusions. First, statistical significance is not operational usefulness: a large corpus can make a weak AUC statistically significant without making it useful for process control. Second, calibration fails: the QE score should not be treated as an absolute quality probability or as a stable translation-quality score. Third, threshold-based triage is fragile: thresholds that catch most errors usually flag most of the file, whereas thresholds that save on review effort miss many errors. Fourth, the behavior of the model is not portable: the same language pair can behave differently across EC and EP datasets, and the same slice can receive different workflow verdicts under different threshold policies.

\paragraph{Production implication.}
The operational implication is that a QE tool such as TAUS EPIC QE cannot be used as a standalone routing, release, or review-bypass mechanism on a new file merely because it produced useful statistics elsewhere. Before using it in production, local validation on the material with the same language pair, domain, MT or AI translation source, editorial condition, and comparable expected error rate would be needed. A defensible workflow would sort segments worst-first, retain random or stratified sampling from the supposedly good tail, and validate against human MQM annotation that the top score bands actually contain a materially higher error density than random selection. Without such validation, the score is not a safe decision variable.

The empirical evidence supports only a qualified, restricted conclusion: TAUS EPIC QE exhibits a modest ranking signal in some language-, domain-, and error-rate regimes, but does not provide a reliable calibrated quality score and does not support unvalidated triage decisions.

This empirical result is consistent with the theoretical critique referenced and developed in the preceding sections. Segment-level QE may contain linguistic or distributional information learned from training data, but this information is not equivalent to information about translation quality. It is a corpus-dependent proxy signal. Consequently, QE outputs should be validated through a rigorous statistical comparison with human quality annotations before being incorporated into workflow decisions. What matters is not the score itself, but whether that score is discriminative, calibrated, threshold-stable, economically meaningful, and safe under the specific conditions of deployment.

\section{Conclusion}

The cumulative evidence drawn from empirical studies, translation quality evaluation research, and the present analysis 
suggests that QE technology continues to exhibit instability across multiple dimensions.
This limitation raises serious concerns about its suitability for real-world applications.

The persistent failure of QE systems to handle linguistic phenomena such as homophones illustrates that current QE systems do not appear to capture language-agnostic indicators of translation quality in a stable manner. The practical utility of QE in professional workflows is also far from established. The findings on length bias further suggest that current architectures often rely on superficial statistical patterns rather than genuine semantic evaluation.

Taken together, domain generalization failures, systematic biases, annotation inconsistencies, and challenges in fine-grained error classification suggest that QE systems still lack a robust and portable understanding of translation quality. More data and more sophisticated architectures have not yet resolved these structural issues.

Future progress toward robust evaluation technology will require explicit attention to the statistical uncertainty inherent in both human and AI judgment. As \newcite{sarti2025annotators} demonstrate, relying on single-pass evaluations yields unstable metric rankings. Regardless of whether quality assessment is performed by human annotators, automated models, or hybrid systems, evaluator variance must be explicitly recognized, measured, and reported. Accounting for this uncertainty—whether through multi-evaluator sampling, prior uncertainty calibration, or probabilistic scoring—is essential for establishing reliable translation quality metrics.

MQM-based approaches differ from QE in that they explicitly model multiple dimensions of translation quality and incorporate structured human judgment, thereby allowing contextual factors to be taken into account. This makes them better suited to capturing the multi-faceted and context-dependent nature of translation quality.
MQM is not free from subjectivity or annotation uncertainty; however, unlike segment-level QE, it makes these sources of uncertainty explicit through structured error taxonomies and human evaluation procedures.

Rather than continuing to prioritize fully automated QE as a standalone approach, future research may benefit from methods that more explicitly account for the complexity of human language and judgment. The available evidence suggests that segment-level QE scores are not sufficiently reliable to serve as standalone signals for routing decisions, release criteria, or the prioritization of human review. Addressing these limitations may require moving beyond isolated, segment-level prediction toward evaluation frameworks grounded in structured, multi-dimensional assessment, such as MQM, potentially supported by multi-evaluator and AI-assisted approaches.

\section*{Disclaimer}
The views expressed in this paper are solely those of the authors and do not represent the official position of the European Parliament or its Directorate-General for Translation and Clear Language.

\section*{Acknowledgments}
We thank Fred Bane, Director of Data Science at TransPerfect for valuable comments on this paper.

\bibliography{references}

\end{document}